\documentclass[sigconf]{acmart}
\usepackage{balance}
\usepackage{graphicx}
\usepackage{booktabs}
\usepackage{algorithm}
\usepackage{algorithmic}
\usepackage[switch]{lineno}

\usepackage{amsthm,amsmath,amssymb}
\usepackage{mathrsfs}
\usepackage{subfigure}
\usepackage{booktabs}
\usepackage{multirow}
\usepackage{hyperref}
\usepackage{multicol}

\AtBeginDocument{%
 }

\copyrightyear{2023}
\acmYear{2023}
\setcopyright{rightsretained}
\acmConference[CIKM '23]{Proceedings of the 32nd ACM International Conference on Information and Knowledge Management}{October 21--25, 2023}{Birmingham, United Kingdom}
\acmBooktitle{Proceedings of the 32nd ACM International Conference on Information and Knowledge Management (CIKM '23), October 21--25, 2023, Birmingham, United Kingdom}\acmDOI{10.1145/3583780.3614804}
\acmISBN{979-8-4007-0124-5/23/10}

\newtheorem{myDef}{Definition}
\begin{document}
% \balance
\title{Causality and Independence Enhancement for Biased Node Classification}

% \author{Anonymous authors}
%  \affiliation{
%    \institution{Paper under double-blind review}
%  }
% \renewcommand{\shortauthors}{Anonymous Author, et al.}
% \affiliation{%

\author{Guoxin Chen}
\affiliation{%
  \institution{Data Intelligence System Research Center, Institute of Computing Technology, Chinese Academy of Sciences}
  \institution{University of Chinese Academy of Sciences}
  % \streetaddress{1 Th{\o}rv{\"a}ld Circle}
  \city{BeiJing}
  \country{China}}
\email{chenguoxin22s@ict.ac.cn}

\author{Yongqing Wang}
\authornote{Corresponding authors: Yongqing Wang and Fangda Guo}
\affiliation{%
  \institution{Data Intelligence System Research Center, Institute of Computing Technology, Chinese Academy of Sciences}
  % \streetaddress{1 Th{\o}rv{\"a}ld Circle}
  \city{BeiJing}
  \country{China}}
\email{wangyongqing@ict.ac.cn}

\author{Fangda Guo}
% \authornote{Corresponding author}
\authornotemark[1]
\affiliation{%
  \institution{Data Intelligence System Research Center, Institute of Computing Technology, Chinese Academy of Sciences}
  % \streetaddress{1 Th{\o}rv{\"a}ld Circle}
  \city{BeiJing}
  \country{China}}
\email{guofangda@ict.ac.cn}

\author{Qinglang Guo}
% \authornote{Corresponding author}
\affiliation{%
  \institution{School of Cyber Science and Technology, University of Science and Technology of China}
  \institution{National Engineering Research Center for Public Safety Risk Perception and Control by Big Data (RPP), China Academic of Electronics and Information Technology}
  % \streetaddress{1 Th{\o}rv{\"a}ld Circle}
  \city{HeFei}
  \country{China}}
\email{gql1993@mail.ustc.edu.cn}

\author{Jiangli Shao}
\affiliation{%
  \institution{Data Intelligence System Research Center, Institute of Computing Technology, Chinese Academy of Sciences}
  \institution{University of Chinese Academy of Sciences}
  % \streetaddress{1 Th{\o}rv{\"a}ld Circle}
  \city{BeiJing}
  \country{China}}
\email{shaojiangli19z@ict.ac.cn}

\author{Huawei Shen}
\author{Xueqi Cheng}
\affiliation{%
  \institution{Data Intelligence System Research Center, Institute of Computing Technology, Chinese Academy of Sciences}
  \institution{University of Chinese Academy of Sciences}
  % \streetaddress{1 Th{\o}rv{\"a}ld Circle}
  \city{BeiJing}
  \country{China}}
\email{shenhuawei@ict.ac.cn}
\email{cxq@ict.ac.cn}

\renewcommand{\shortauthors}{Guoxin Chen et al.}
%% use of ampersand (\&) versus ''and'' is to saves space.

% Recently, there has been an increasing interest in the problem of out-of-distribution (OOD) generalization on graphs.
\begin{abstract}
Most existing methods that address out-of-distribution (OOD) generalization for node classification on graphs primarily focus on a specific type of data biases, such as label selection bias or structural bias. However, anticipating the type of bias in advance is extremely challenging, and designing models solely for one specific type may not necessarily improve overall generalization performance. Moreover, limited research has focused on the impact of mixed biases, which are more prevalent and demanding in real-world scenarios. To address these limitations, we propose a novel Causality and Independence Enhancement (CIE) framework, applicable to various graph neural networks (GNNs). Our approach estimates causal and spurious features at the node representation level and mitigates the influence of spurious correlations through the backdoor adjustment. Meanwhile, independence constraint is introduced to improve the discriminability and stability of causal and spurious features in complex biased environments. Essentially, CIE eliminates different types of data biases from a unified perspective, without the need to design separate methods for each bias as before. To evaluate the performance under specific types of data biases, mixed biases, and low-resource scenarios, we conducted comprehensive experiments on five publicly available datasets. Experimental results demonstrate that our approach CIE not only significantly enhances the performance of GNNs but outperforms state-of-the-art debiased node classification methods.
\end{abstract}

\begin{CCSXML}
<ccs2012>
   <concept>
       <concept_id>10010147.10010257</concept_id>
       <concept_desc>Computing methodologies~Machine learning</concept_desc>
       <concept_significance>500</concept_significance>
       </concept>
 </ccs2012>
\end{CCSXML}

\ccsdesc[500]{Computing methodologies~Machine learning}

\keywords{Node classification; Out-of-distribution generalization; Graph neural networks}



\maketitle

\section{Introduction}
Recently, there has been increasing interest in the problem of out-of-distribution (OOD) generalization for node classification.
On the one hand, although traditional graph neural networks \cite{wu2020comprehensive,defferrard2016convolutional,kipf2016semi,velivckovic2017graph} have achieved tremendous success in node classification problems\cite{zhuang2022robust,choi2022finding}, they are limited by the independent and identically distributed (I.I.D.) assumption. 
On the other hand, unlike the image or text fields\cite{ahmed2020systematic,ren2022out,rosenfeld2020risks}, the non-Euclidean space of graph structures poses unique challenges for the development of OOD generalization algorithms on graphs\cite{li2022out,chen2022indni}.
Moreover, distribution shift on graphs can exist at the feature level (e.g., node features) and the topology level (e.g., graph structural properties), which may arise from different types of data biases respectively\cite{li2022out}, rendering more challenges for OOD generalization on graphs.

Prior research has focused on designing methodologies for a specific type of data biases in node classification.
For example, to address the structural bias (which results in topology-level distribution shift), CATs\cite{he2021learning} consider structural perturbations during training and propose a joint attention mechanism that adaptively weights the aggregation of each node, thereby enhancing the robustness.
To address the label selection bias (which results in feature-level distribution shift), DGNN \cite{fan2022debiased} incorporates the idea of decorrelating variables from causal learning into GNNs, reducing the impact of bias on the model.
While these methods have achieved certain levels of success, but they have been purposefully designed to target only one specific type of biases.
However, in reality, it is challenging to anticipate the type of bias in advance, and designing methods targeting one specific type alone may not necessarily improve the generalization performance overall.
From subsequent experiments in the Section \ref{section:main_exp}, it can be observed that some methods even sacrifice the robustness of models to the other types of data biases in order to improve the performance on the specific one.

Recently, EERM\cite{wu2022handling} has taken into account different types of data biases from the perspective of IRM\cite{arjovsky2019invariant,rosenfeld2020risks}. It attempts to learn invariant features by minimizing the mean and variance of risks across multiple environments simulated by the adversarial contextual generator.
However, unlike image or text data \cite{ren2022out,ahmed2020systematic,gao2023out} where multiple environments naturally exist, EERM requires sufficient annotation information to generate high-quality environments through an adversarial contextual generator; otherwise, it may introduce more noise into the model.
Furthermore, data is often affected by multiple types of biases simultaneously (which we refer to as \textbf{mixed biases}).
Generating high-quality environments in complex biased contexts, especially in mixed biases, presents a significant challenge. To the best of our knowledge, limited research has investigated the effects of mixed biases, despite their evident prevalence and increasing demand.

To address the aforementioned issues, we propose a \textbf{C}ausality and \textbf{I}ndependence \textbf{E}nhancement framework (CIE) to improve the generalization of traditional graph neural networks against different types of data biases, especially mixed biases\footnote{The code for this paper can be found at \url{https://github.com/Chen-GX/CIE}}.
Intuitively, within the framework of graph neural networks, different types of biases are incorporated into the node representations.
Our proposed method CIE takes a unified perspective to simultaneously debias different types of data biases at the node representation level, overcoming the limitation of designing methods for each type of biases separately.
% Applying a unified strategy to remove different types of biases at the node representation level is a more effective and efficient approach as it enables the model to handle multiple types of biases simultaneously.
% Intuitively, within the framework of graph neural networks, a more effective and efficient approach is to incorporate different types of biases into the node representations, enabling the model to handle them simultaneously.
However, our analysis of causal graphs in Section \ref{section:Causal_perspective} reveals that traditional graph neural networks fail to differentiate stable causal features from spurious features that are easily affected by the environment\cite{chen2022learning,lin2021generative}.
This makes them vulnerable to the impact of data biases, due to the establishment of spurious correlations.
% However, our analysis of causal graphs in Section \ref{section:Causal_perspective} reveals that traditional graph neural networks, due to their reliance on spurious correlations\cite{chen2022learning,lin2021generative}, fail to differentiate stable causal features from spurious features that are easily affected by the environment. This makes them vulnerable to the impact of data biases.
Therefore, we propose to estimate causal and spurious features and mitigate the influence of spurious correlations through the backdoor adjustment\cite{pearl2009causal,bareinboim2016causal} to eliminate the impact of data biases.
Moreover, the complex biased environments (especially mixed biases) pose a challenge to the estimation of causal and spurious features.
Through the analysis of causal graphs (see Section \ref{section:Causal_perspective} for details), we propose the independence constraint to enhance the discriminability between causal and spurious features and improve the stability of estimation in the complex biased environments.
Extensive experimental results confirm the effectiveness of our proposed method, which can be widely applied to graph neural networks, such as GCN\cite{kipf2016semi}, GraphSAGE\cite{hamilton2017inductive}, and GAT\cite{velivckovic2017graph}, and improve the generalization of GNNs against different types of data biases, especially mixed biases.

The contributions of this paper are
summarized below.
\begin{itemize}
    \item We propose a novel Causality and Independence Enhancement (CIE) framework that enhances the generalization of various GNNs. This framework introduces the independence constraint to improve discriminability and stability of causal and spurious features in complex biased environments and  mitigates spurious correlations by implementing the backdoor adjustment.
    \item To the best of our knowledge, this is the first study to analyze the impact of mixed biases in node classification.
    Our proposed CIE integrates different types of biases into a unified perspective, thus enabling the model to simultaneously handle multiple types of biases and eliminating the need for designing separate methods for each bias.
    \item We conducted extensive experiments on five publicly available datasets to comprehensively evaluate the performance under specific types of data biases, mixed biases, and low-resource scenarios. The experimental results demonstrate the effectiveness and efficiency of our proposed method.
\end{itemize}

The remainder of this paper is organized as follows. 
Section \ref{section:related_work} reviewed related work. Section \ref{section:preliminary} formalized the problem and analyzed node classification from the causal perspective. Section \ref{section:method} provided a detailed exposition of the proposed CIE method. Section \ref{section:experiments} conducted extensive experiments on various publicly available graph benchmarks.
Section \ref{section:conclusion} presented the conclusion of this paper.

\section{Related Work}
\label{section:related_work}
The goal of out-of-distribution (OOD) generalization\cite{shen2021towards,ye2021towards,li2022out, hendrycks2016baseline,liu2020energy} is to enhance the model's generalization performance and robustness in the face of unknown distribution shifts.
Since most real-world applications do not satisfy the independent and identically distributed (I.I.D.) assumption, the OOD generalization problem has always occupied an important position.
Different from the data such as images and texts \cite{ahmed2020systematic,ren2022out}, the non-Euclidean nature of graph structure hinders the development of OOD generalization algorithms on graphs.
Furthermore, the presence of complex types of graph distribution shift, such as feature-level and topology-level distribution shifts, renders OOD generalization on graphs even more challenging.

\textbf{Debiased Graph classification.}
Recently, there have some attempts\cite{li2022graphde,li2022learning,yang2022learning,wang2022imbalanced} to solve the OOD generalization problem on graph classification.
For example, GraphDE\cite{li2022graphde} redefines the correlation between debias learning and OOD detection in a probabilistic framework and proposes an OOD detector to identify outliers and reduce their weights for debiasing.
CAL\cite{sui2022causal} proposes a causal attention mechanism for graph classification that aims to filter out shortcut patterns and improve generalization.
MRL\cite{yang2022learning} designs new objective functions based on the principle of environmental invariance of specific substructures in graph classification to learn invariant features for debiasing purposes.
Different from node classification, the graph classification problem analyzes and processes entire graphs by various techniques, focusing more on extracting core subgraphs.

\textbf{Debiased Node classification.}
At present, there are some research studies\cite{tang2020investigating,qu2021imgagn,fan2022debiased,wu2022handling} have been conducted to explore methods for mitigating specific types of data biases in node classification problems.
For example, 
CATs\cite{he2021learning} propose a joint attention that considers structural perturbations to improve the generalization of the model to the structural bias.
BA-GNN\cite{chen2022ba} improves the generalization of the model by recognizing biases and learning invariant features.
DGNN\cite{fan2022debiased} introduces the idea of variable decorrelation into GNNs to alleviate the influence of label selection bias.
However, designing models solely for specific type of biases does not genuinely enhance generalization due to the challenge of predicting the exact types of data biases in practical applications.
Although EERM\cite{wu2022handling} considers different types of data biases, it heavily relies on sufficient annotation information to generate multiple high-quality environments.
Moreover, limited research has explored the impact of mixed biases, which are more prevalent and demanding in real-world applications.
In this paper, our approach takes a unified strategy on different types of data biases that may result in feature-level or topology-level distribution shift.
Similar to CAL, we take a causal perspective and learn causal features while removing spurious correlations.
However, CAL does not consider the relationship between causal features and spurious features, which is fatal at the more detail-oriented node level.

\section{Preliminary} 
\label{section:preliminary}
% We first analyze the impact of biases on node classification from a causal perspective and then provide a detailed introduction to the proposed method and its implications.
\subsection{Problem Formulation}
We provide a formal definition of the OOD problem in node classification.
Let $\mathbb{X}$ be the feature space of nodes (including topological features), and $\mathbb{Y}$ be the label space. A node classifier $f_{\theta}$ maps input instances $x\in \mathbb{X}$ to the label $y\in \mathbb{Y}$.
\begin{myDef}[The OOD problem of node classification]
Given a training set $D=\{(x_i,y_i)\}_{i=1}^{N}$ of $N$ nodes sampled from the training distribution $P_{train}(X,Y)$, where $x_i \in \mathbb{X}$ and $y_i \in \mathbb{Y}$, the goal is to train an optimal classifier $f_{\theta}^{*}$ to achieve optimal generalization on the test data sampled from the testing distribution $P_{test}(X, Y)$, where $P_{train}(X,Y) \neq P_{test}(X, Y)$:
\begin{equation}
    f_{\theta}^{*} = \mathop{\arg\min}\limits_{f_{\theta}}\mathbb{E}_{X,Y\sim P_{test}}[\ell(f_{\theta}(X), Y)],
\end{equation}
where $\ell$ is the loss function.
\end{myDef}
Directly minimizing the training loss on $\mathbb{E}_{X,Y\sim P_{train}}[\ell(f_{\theta}(X), Y]$ can not obtain the optimal classifier $f_{\theta}^{*}$ on the test set, because the distribution shift between $P_{train}$ and $P_{test}$ breaks the I.I.D. assumption under different types of data biases.
However, the causal relationship between causal features and labels is not affected by distribution shift \cite{pearl2000models,chen2022learning}.
Therefore, in this paper, we propose the CIE framework from a causal perspective to steer the model to establish $f_{\theta}^{\dag}$ between causal features and labels to achieve optimal results in $P_{test}$.

\subsection{Causal perspective in node classification}\label{section:Causal_perspective}
\begin{figure}[]
    \centering
    \includegraphics[width=\linewidth]{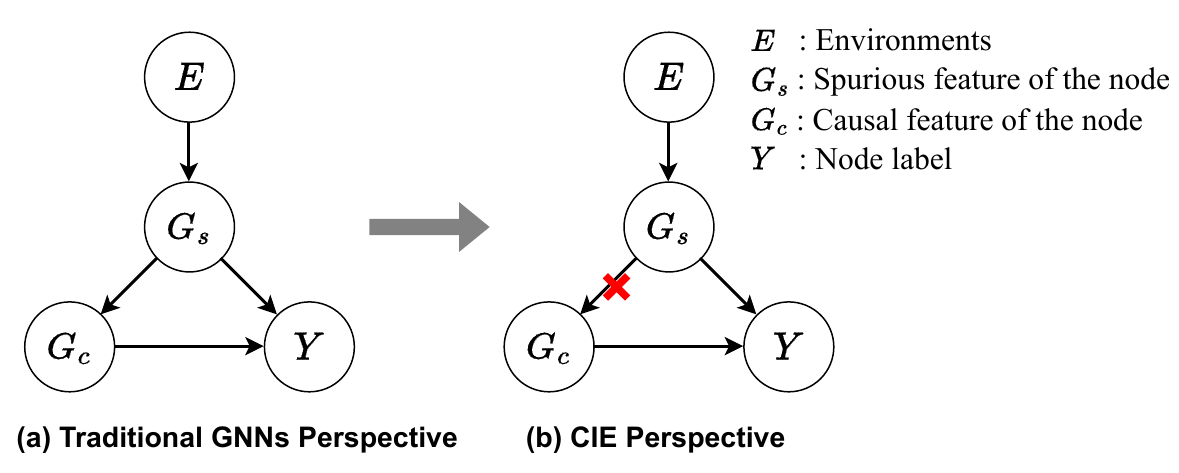}
    \caption{Causal graph on Node Classification. (a) Node classification Process from the perspective of traditional GNNs. (b) Node Classification Process Revised by CIE.}
    \label{fig:SCM}
\end{figure}
We revisit the node classification process of traditional graph neural networks from the perspective of causal learning\cite{pearl2000models,spirtes2010introduction}, and assume that the nodes in the graph are generated through a mapping $f_{gen}:\mathcal{Z} \rightarrow\mathcal{G}$, where $\mathcal{Z}\subseteq \mathbb{R}^{n}$ is the latent space and $\mathcal{G}$ is the graph space. Following the previous work\cite{von2021self,chen2022learning}, we divide the latent variable $\mathcal{Z}$ into spurious features $G_s$ and causal features $G_c$ according to whether it is directly affected by the environment $E$. 

In theory, $G_s$ and $G_c$ are two independent subspaces of $\mathcal{Z}$ \cite{von2021self,chen2022learning}. 
However, the traditional graph neural networks focus on establishing the correlation between node features and labels under the independent and identically distributed assumption, while ignoring the fact that spurious features $G_s$ are easily affected by the environment $E$.
Therefore, the node representations learned by the traditional graph neural networks are the fusion of $G_s$ and $G_c$. $G_s$ confuses the causation between causal features $G_c$ and labels $Y$ in the prediction stage, $Y:=f_{gnns}(G_s, G_c)$, as illustrated in Fig.\ref{fig:SCM}(a).
Moreover, some recent work\cite{torralba2011unbiased,geirhos2020shortcut,lin2021generative} has found that neural networks, including graph neural networks, are more susceptible to establishing spurious correlations\cite{chen2020self,zhou2021examining,khani2021removing} between spurious features and labels in data, which are not stable in nature.

Therefore, in this paper, we aim to address the limitations of traditional graph neural networks from a causal perspective. Specifically, we treat $G_s$ as a confounder on the backdoor path\cite{pearl2009causal,spirtes2010introduction} ($G_c\leftarrow G_s\rightarrow Y$) between $G_c$ and $Y$.
Theoretically, we can block the above backdoor path through the backdoor adjustment\cite{pearl2009causal,spirtes2010introduction}, since $G_s$ satisfies the backdoor criterion\cite{bareinboim2016causal}, as illustrated in Fig.\ref{fig:SCM}(b). 
Formally, we can introduce the following formula through the backdoor adjustment:
\begin{equation}
    \begin{aligned}
        P(Y|\operatorname{do}(G_c)) 
        & = \sum_{s\in G_s} P(Y | \operatorname{do}(G_c), s) P(s|\operatorname{do}(G_c))\\
        & = \sum_{s\in G_s} P(Y|G_c,s)P(s|\operatorname{do}(G_c)) \\
        & = \sum_{s\in G_s} P(Y|G_c, s)P(s),
    \end{aligned}
    \label{eq:back_adjustment}
\end{equation}
where $P(Y|G_c, s)$ represents the conditional probability of $Y$ given the causal feature $G_c$ and the confounder $s$, and $P(s)$ is the prior probability of the confounder.

However, the aforementioned formula presents two challenges: (1) Due to the characteristics of graph data, $G_s$ is typically not observable, making it difficult to directly obtain $G_s$ and $G_c$ at the data level. (2) Given that $G_s$ is a continuous subspace of $\mathcal{Z}$, the process of discretely integrating it is extremely challenging, as shown in Equation \ref{eq:back_adjustment}.
In the rest of this paper, we dedicate to elaborating on the methodology to overcome these challenges and efficiently approximate Equation \ref{eq:back_adjustment}, demonstrating the efficiency and effectiveness.

\section{METHODOLOGY}\label{section:method}
In this section, we first introduce the overall framework of the Causality and Independence Enhancement (CIE), followed by a detailed description of the proposed method.

\subsection{Overall Framework}
\begin{figure}[]
    \centering
    \includegraphics[width=0.85\linewidth]{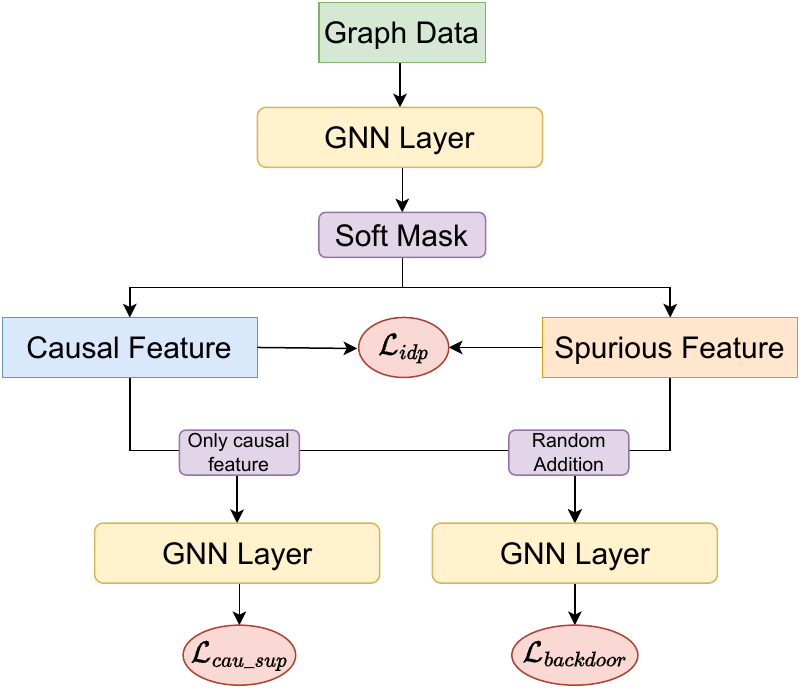}
    \caption{The overall framework of CIE.}
    \label{fig:framework}
\end{figure}

The structure of the proposed CIE framework is illustrated in Figure \ref{fig:framework}, which consists of two main steps: Estimating causal and spurious features from node representations, and Mitigating spurious correlations.
Specifically, causal and spurious features are estimated from node representations by the soft mask module as well as independence constraint.
In addition, spurious correlations are mitigated by blocking the backdoor path ($G_c\leftarrow G_s\rightarrow Y$) through the backdoor adjustment (approximating Equation \ref{eq:back_adjustment}).

In previous works, methods designed to address a specific type of data bias may not necessarily enhance generalization performance overall, due to the lack of global considerations.
Furthermore, we found that when labeled information is insufficient, the prediction of graph structure in EERM could introduce fatal misinformation to the nodes with lower degree.
Therefore, our approach debiases different types of data biases from a unified perspective, avoiding the need to design methods for each bias separately and preventing the introduction of misinformation in low-resource scenarios.
Our intuition is straightforward that either feature-level bias or topology-level bias will ultimately be incorporated into the node representations in the graph neural network framework.
From a causal perspective, identifying and eliminating data biases in node representations through the backdoor adjustment is a more effective and efficient approach.
Moreover, the complex biased environments (especially mixed biases) makes it more difficult to disentangle the spurious features, which poses a challenge for the estimation of causal and spurious features.
The introduction of independence constraint not only improves the discriminability between causal and spurious features but also enhances stability in complex biased environments.
Subsequent experimental results have also confirmed our hypotheses.

Next, we provide a detailed description of the steps involved in estimating causal and spurious features and mitigating spurious correlations.

\subsection{Disentanglement from Node Representation}
As previously mentioned, identifying causal and spurious features ($G_c$ and $G_s$, respectively) at the graph data level is often challenging due to the discrete nature of graph data.
However, traditional graph neural networks are capable of learning representations for each node that incorporate causal and spurious features, which provides an opportunity to address the aforementioned challenge by disentangling these two types of features at the node representation level.

We introduce the soft mask module to generate two branches ($\mathbf{H_c}$ and $\mathbf{H_s}$) for estimating causal features $G_c$ and spurious features $G_s$, respectively.
The soft mask module consists of $\mathbf{M_c}\in \mathbb{R}^{N\times 1}$, $\mathbf{M_s} \in \mathbb{R}^{N\times 1}$ and $\mathbf{M_c} + \mathbf{M_s} = \mathbf{1}$, where $N$ is the number of nodes and $\mathbf{1}$ is the all-one matrix.

To extract features and obtain node representations $\mathbf{H}$, we use a GNN layer to aggregate neighbor information from the graph $G=\{\mathbf{A},\mathbf{X}\}$, where $\mathbf{A}$ is the adjacency matrix and $\mathbf{X}$ is the node feature matrix. The resulting node representations $\mathbf{H}$ incorporates both spurious and causal features, as discussed in Section \ref{section:Causal_perspective}.
\begin{equation}
    \mathbf{H} = \operatorname{GNN}(\mathbf{A},\mathbf{X}).
    % \nonumber
\end{equation}
The weight of the two branches is estimated by a multi-layer perceptron (MLP), which determines the value of the soft mask.
\begin{equation}
    \centering
    \begin{aligned}
        \mathbf{M_c}, \mathbf{M_s} &= \sigma(\operatorname{MLP}(\mathbf{H})) \\
        \mathbf{H_c} = \mathbf{H} \odot &\mathbf{M_c}, \; \mathbf{H_s} = \mathbf{H} \odot \mathbf{M_s},
    \end{aligned} 
\end{equation}
where $\sigma$ is the softmax function. 
% Although $\mathbf{H_c}$ and $\mathbf{H_s}$ are obtained from node representations containing $G_c$ and $G_s$, it still needs to satisfy the following constraints to serve as approximate estimates of $G_c$ and $G_s$:
$\mathbf{H_c}$ and $\mathbf{H_s}$ are derived from node representations $\mathbf{H}$ containing $G_c$ and $G_s$, but they still need to satisfy the following constraints to serve as approximate estimates of $G_c$ and $G_s$:
\begin{enumerate} %itemize
	\item \textbf{Independence.} $\mathbf{H_c}$ and $\mathbf{H_s}$ should be as independent as possible, as mentioned in Section \ref{section:Causal_perspective} for $G_c$ and $G_s$.
	\item \textbf{Backdoor adjustment.} The backdoor path is broken off by Equation \ref{eq:back_adjustment} to mitigate spurious correlations, as shown in Fig.\ref{fig:SCM}(b).
\end{enumerate}

\subsection{Independence}
According to the causal perspective discussed in Section \ref{section:Causal_perspective}, $G_c$ and $G_s$ are two subspaces of the latent space $\mathcal{Z}$ that are independent of each other.
As approximate estimates of $G_c$ and $G_s$, $\mathbf{H_c}$ and $\mathbf{H_s}$ should also be as independent of each other as possible.
Intuitively, independence provides greater stability and discriminability in the presence of different types of biases, as confirmed by subsequent experimental results.

We define $\mathcal{X}$ and $\mathcal{Y}$ as two spaces of $G_c$ and $G_s$, respectively, and each sampling gets a pair of causal and spurious features of a node $i$, $\{(h_{ci}, h_{si}) \in \mathcal{X} \times \mathcal{Y}\}$. 
Next, we define mapping functions $\phi(h_{ci}) \in \mathcal{F}$ and $\psi(h_{si}) \in \varOmega$ that map each $h_{ci} \in \mathcal{X}$ and $h_{si} \in \mathcal{Y}$ to the kernel space $\mathcal{F}$ and $\varOmega$ with respect to the kernel functions $k_{\phi}(h_{ci}, h_{ci}) =<\phi(h_{ci}), \phi(h_{ci})>$ and $k_{\psi}(h_{si}, h_{si}) =<\psi(h_{si}), \psi(h_{si})>$, respectively. $\mathcal{F}$ and $\varOmega$ are Reproducing Kernel Hilbert Space (RKHS)\cite{paulsen2016introduction} on $\mathcal{X}$ and $\mathcal{Y}$, respectively. According to \cite{baker1973joint,fukumizu2004dimensionality}, we can define a cross-covariance operator between the above feature maps:
\begin{equation}
    \mathcal{C}_{h_{ci}h_{si}} = \mathbf{E}_{h_{ci},h_{si}}[(\phi(h_{ci})-\mu_{h_{ci}}) \otimes (\psi(h_{si})-\mu_{h_{si}})],
\end{equation}
where $\otimes$ is the tensor product, and $\| \mu_{h_{ci}} \|_{\mathcal{F}}^2$ can be calculated by applying the expectation twice via E.q.\ref{eq:mu}, and $\| \mu_{h_{si}} \|_{\varOmega}^2$  similarly.
\begin{equation}
    \| \mu_{h_{ci}} \|_{\mathcal{F}}^2 = \mathbf{E}_{h_{ci},h_{ci}}[k_{\phi}(h_{ci},h_{ci})].
    \label{eq:mu}
\end{equation}

Then, the square of the Hilbert-Schmidt norm of the correlated cross-covariance operator (HSIC), $\mathcal{C}_{\mathbf{H_{c}}\mathbf{H_{s}}}$ is expressed as
\begin{equation}
    \operatorname{HSIC}(\mathcal{F}, \varOmega, P_{\mathbf{H_{c}},\mathbf{H_{s}}})= \|\mathcal{C}_{\mathbf{H_{c}}\mathbf{H_{s}}} \|_{\operatorname{HS}}^2,
\end{equation}
where $P_{\mathbf{H_{c}},\mathbf{H_{s}}}$ is a joint measure over $\mathcal{F} \times \varOmega$ and $\| X\|_{\operatorname{HS}} = \sqrt{\sum_{i,j}x_{i,j}^2}.$
Therefore, the empirical estimate of the HSIC \cite{gretton2005measuring,song2007supervised} is given as follows
\begin{equation}
    \operatorname{HSIC}(\mathcal{F}, \varOmega, Z) = (m-1)^2 \operatorname{tr}(\mathbf{K}\mathbf{R}\mathbf{L}\mathbf{R}),
\end{equation}
where $Z=\{(h_{c1},h_{s1}),...,(h_{cm},h_{sm})\} \subseteq \mathcal{F} \times \varOmega$, $\mathbf{K},\mathbf{L},\mathbf{R}\in \mathbb{R}^{m\times m}$, $\mathbf{K}_{ij}=k_{\phi}(h_{ci},h_{cj})$, $\mathbf{L}_{ij}=k_{\psi}(h_{si},h_{sj})$, and $R$ is the centering matrix $\mathbf{R}_m=\mathbf{I}_m-\frac{1}{m}\mathrm{11^T}$.
The independence between causal and spurious features can be described by HSIC using kernel functions, with the computational complexity of $\mathcal{O}(m^2)$, where m is the number of samples.
In contrast, other kernel methods require at least $\mathcal{O}(m^3)$\cite{bach2002kernel}.
But HSIC is not invariant to isotropic scaling, we adopt the normalized HSIC (CKA) \cite{kornblith2019similarity}, which is defined as follows:
\begin{equation}
    \mathcal{L}_{idp} = \operatorname{CKA}(\mathbf{H_{c}},\mathbf{H_{s}}) = \frac{\operatorname{HSIC}(\mathbf{H_{c}},\mathbf{H_{s}})}{\sqrt{\operatorname{HSIC}(\mathbf{H_{c}},\mathbf{H_{c}}) \operatorname{HSIC}(\mathbf{H_{s}},\mathbf{H_{s}})}}.
    \label{eq:idp}
\end{equation}

CKA describes the correlation between the causal and spurious feature subspaces. When $k_{\phi}$ and $k_{\psi}$ are universal kernels, $\operatorname{CKA}=0$ implies independence.
We aim to minimize the dependency between $\mathbf{H_c}$ and $\mathbf{H_s}$ by introducing the loss as depicted in Equation \ref{eq:idp}. In the experimental section, we focus on employing radial basis functions\cite{buhmann2003radial} and investigate the impact of different kernel functions.

\subsection{Backdoor Adjustment}
Traditional graph neural networks often suffer from unstable spurious correlations between $G_c$ and $Y$, caused by the presence of the backdoor path ($G_c\leftarrow G_s\rightarrow Y$), as illustrated in Fig.\ref{fig:SCM}.
It is evident that when the data does not follow the I.I.D. assumption, the spurious correlation between $G_c$ and $Y$ can lead to decreased performance in graph neural networks.

Our objective is to enhance the generalization performance of GNNs in the presence of different types of biases by applying the backdoor adjustment to eliminate spurious correlations.
However, there exists a challenge when attempting to discretize $G_s$(or $H_s$) as described in Section \ref{section:Causal_perspective}.
Inspired by \cite{pearl2014interpretation,sui2022causal}, we intervene on the backdoor path at the node representation level and introduce random addition to approximate Equation \ref{eq:back_adjustment}.
Specifically, in accordance with the principles of backdoor adjustment, we hereby introduce the following loss function:

\begin{equation}
    \begin{aligned}
    z_{i}^{\prime} = \operatorname{GNN}(\mathbf{A}, &h_{ci} + h_{sj}), \; sj \in \mathbf{H_s},
    % \hat{y}_{bki} &= \sigma(z_{i}^{\prime})      
    \end{aligned}
\end{equation}
\begin{equation}
    \mathcal{L}_{backdoor} = \frac{1}{n \times |\mathbf{H_s}|} \sum_{i=1}^n \sum_{sj\in \mathbf{H_s}} \ell (y_i, \sigma(z_{i}^{\prime})),
\end{equation}
where $h_{ci}$ is the causal feature of node $i$ from $\mathbf{H_s}$, $h_{sj}$ is the spurious feature of node $j$ randomly sampled from $\mathbf{H_s}$, $z_i^{\prime}$ is the node representations after intervention, $\sigma$ is the Softmax function, $n$ is the number of training examples, and $\ell$ is the cross-entropy loss.
Specifically, we assume that $P(s)$ follows a uniform distribution, and estimate $P(Y|G_c, s)$ by $\frac{1}{n} \sum_{i=1}^n \ell (y_i, \sigma(z_{i}^{\prime}))$ through randomly sampling from $\mathbf{H_s}$ ($\mathbf{H_s}$ is the approximate estimate of $G_s$).

Thus far, we have successfully disentangled $G_c$ and $G_s$ at the node representation level and eliminated the influence of the backdoor path between $G_c$ and $Y$, resulting in causal features for each node. 
We use the message passing mechanism to aggregate and propagate the causal features of neighboring nodes, resulting in node representations for the final classification objective.

\begin{equation}
    z_{ci} = \operatorname{GNN}(\mathbf{A},\mathbf{H_c}),
    % \hat{y_c} = \sigma(z_{ci})
    % \nonumber
\end{equation}
\begin{equation}
    \mathcal{L}_{cau\_sup} = \frac{1}{n} \sum_{i=0}^{n} \ell (y_i, \sigma(z_{ci})),
    \label{eq:sup}
\end{equation}
where $z_{ci}$ is the representation for the i-th node, $n$ is the number of training examples.

In summary, the overview of CIE is shown in Figure \ref{fig:framework}, and the objective function of the CIE consists of the weighted sum of the above-mentioned losses:
\begin{equation}
    \mathcal{L} = \mathcal{L}_{cau\_sup} + \lambda_1 \mathcal{L}_{idp} + \lambda_2 \mathcal{L}_{backdoor}
\end{equation}
where $\lambda_1$ and $\lambda_2$ are the hyperparameter weights that control each component.

% \begin{table*}[]
% \centering
% \caption{Dataset statistics}
% \resizebox{.8\textwidth}{!}{
%     \begin{tabular}{@{}cccccccc@{}}
%     \toprule
%     \textbf{Dataset} & \textbf{Type}    & \textbf{Nodes} & \textbf{Edges} & \textbf{Features} & \textbf{Classes} & \textbf{Bias Type}                & \textbf{Bias degree}             \\ \midrule
%     Cora             & Citation Network & 2,708          & 10556          & 1433              & 7                & Label selection / Structural bias & unbiased/0.4/0.6/0.8             \\
%     Citeseer         & Citation Network & 3327           & 9104           & 3703              & 6                & Label selection / Structural bias & unbiased/0.4/0.6/0.8             \\
%     Pubmed           & Citation Network & 19717          & 88648          & 500               & 3                & Label selection / Structural bias & unbiased/0.4/0.6/0.8             \\
%     NELL             & Knowledge Graph  & 65755          & 251550         & 61278             & 186              & small sample selection bias                 & 1/3/5/10 labeled nodes per class \\
%     Reddit           & Social Network   & 232965         & 114615892      & 602               & 41               & small sample selection bias                 & 1/3/5/10 labeled nodes per class \\ \bottomrule
%     \end{tabular}
%     }
% \label{tab:dataset}
% \end{table*}

\begin{table}[]
\centering
\caption{Dataset statistics}
\resizebox{.98\linewidth}{!}{
\begin{tabular}{cccccc}
\hline
\textbf{Dataset} & \textbf{Type}    & \textbf{Nodes} & \textbf{Edges} & \textbf{Features} & \textbf{Classes} \\ \hline
Cora             & Citation Network & 2,708          & 10556          & 1433              & 7                \\
Citeseer         & Citation Network & 3327           & 9104           & 3703              & 6                \\
Pubmed           & Citation Network & 19717          & 88648          & 500               & 3                \\
NELL             & Knowledge Graph  & 65755          & 251550         & 61278             & 186              \\
Reddit           & Social Network   & 232965         & 114615892      & 602               & 41               \\ \hline
\end{tabular}
    }
\label{tab:dataset}
\end{table}

\section{Experiments}\label{section:experiments}
In this section, we provide a detailed description of the experimental settings and evaluate  the efficiency and effectiveness of our proposed method. The experiments include node classification with different types of data biases, ablation studies, parameter sensitivity analyses, and time complexity comparisons, all aimed at answering the following research questions (RQs):

\begin{itemize}
    \item \textbf{RQ1:} How does CIE compare to state-of-the-art debiased node classification methods in different types of data biases? Additionally, how much improvement can it bring to the backbone models?
    \item \textbf{RQ2:} What is the impact of applying the independence constraint and backdoor adjustment on the proposed model? Can either loss function be effective when used alone?
    \item \textbf{RQ3:} How do different hyperparameters (e.g., the choice of $\lambda_1$ and $\lambda_2$) affect the performance of CIE? Additionally, how does the time efficiency of the proposed method compare to state-of-the-art methods?
\end{itemize}

\subsection{Experimental Settings} \label{section:exp_setting}
To satisfy the definition of OOD, we follow the inductive setting in \cite{wu2019simplifying,fan2022debiased}, masking out test nodes as $G_{train}$ during the training and validation stages, and subsequently predicting the labels of test nodes with the complete graph $G_{test}$. 
In this way, the test distribution is guaranteed to be agnostic during the training phase. 
Moreover, all of the following biases are introduced in the training graph $G_{train}$ to ensure that $P_{train}(X,Y) \neq P_{test}(X,Y)$.

\subsubsection{Different biases in node classification} \label{section:bias}
Currently, most existing works focus on developing algorithms for one specific type of data bias. However, this approach does not necessarily improve generalization overall, since predicting the type of data biases can be difficult in reality.
Therefore, in this paper, we consider several common biases that have been explored in prior research, including \textbf{label selection bias}\cite{zadrozny2004learning,fan2022debiased} that can cause distribution shift at the feature level, \textbf{structural bias}\cite{he2021learning} that can cause distribution shift at the topology level, and their mixed biases.
Additionally, we introduce the \textbf{small sample selection bias}\cite{fan2022debiased} to evaluate the performance of CIE in low-resource scenarios, which can cause distribution shift at the feature level.

\subsubsection{Datasets and Bias Setting}
To evaluate the performance of our proposed method under specific bias, we introduce label selection bias and structural bias on the Cora, Citeseer, and Pubmed datasets\cite{sen2008collective}.
Following \cite{fan2022debiased}, we regulate the level of data bias by adjusting the hyperparameter $\epsilon \in \{0.4, 0.6, 0.8\}$.
Specifically, in the case of label selection bias, a higher value of $\epsilon$ indicates a greater tendency to select nodes with inconsistent labels with their neighboring nodes as the training set.
In the case of structural bias, the value of $\epsilon$ represents the percentage of edges to be randomly removed from $G_{train}$. For example, $\epsilon=0.4$ indicates randomly deleting $40\%$ of the edges.
We select 20 nodes from each class as the training nodes and use the training set partitioned in the original paper\cite{kipf2016semi,velivckovic2017graph} as the \emph{unbiased} setting.

To evaluate performance under mixed biases, we primarily introduce different degrees of label selection bias and structural bias simultaneously on three citation network datasets.
We utilize the hyperparameter $\epsilon$ mentioned above to regulate the combination of these two biases at varying levels.

To evaluate performance in low-resource scenarios, we introduce small sample selection bias on NELL\cite{glorot2010understanding} and Reddit datasets\cite{hamilton2017inductive}.
Specifically, we randomly select 1/3/5/10 nodes from each class as the training set. 
The validation set and test set remain the same as in the original paper\cite{velivckovic2017graph,hamilton2017inductive}.
Table \ref{tab:dataset} displays five publicly available datasets, along with their corresponding bias configurations.

\subsubsection{Baselines}
In order to verify the effectiveness of the proposed method CIE, we conducted a comprehensive comparison with the following eight state-of-the-art methods, which can be classified into three categories:
(1) \textbf{Backbone models}, such as GCN \cite{kipf2016semi}, GraphSAGE \cite{hamilton2017inductive}, and GAT \cite{velivckovic2017graph}. Our method can be directly applied to backbone models to improve their generalization. The improvement in the performance of backbone models can directly validate the effectiveness of the proposed method.
(2) \textbf{Node classification models}. In addition to backbone networks, there have been outstanding works such as SGC\cite{wu2019simplifying} and APPNP\cite{klicpera2018predict} aimed at improving the performance of node classification.
(3) \textbf{Debiased node classification models}. 
Methods tailored to address a specific type of data bias, including CATs\cite{he2021learning}, DGNN\cite{fan2022debiased}, and EERM\cite{wu2022handling}, represent strong baselines in this field.
For DGNN, we compared two variants: DGCN and DGAT.
As for EERM, we utilized the best performing version based on GAT.  

\subsubsection{Implementation Details} 
We implemented our proposed model based on PyTorch\footnote{\url{https://pytorch.org/}} and PyTorch Geometric\footnote{\url{https://www.pyg.org/}}.
% and have made the code publicly available on GitHub\footnote{https://github.com/xxxxxxxx}
For all baselines, all hyperparameter settings are based on the reports in the original papers to achieve optimal results.
For our proposed model, in order to make a fair comparison, we adopt a two-layer architecture for GNNs, where we sample 25 and 10 neighbors separately if it involves the SAGE layer.
We use Adam as the optimizer, with a learning rate of 0.01. The hyperparameters $\lambda_1$ and $\lambda_2$ are tuned in the range of $\{0.1, 0.2, 0.3, ..., 0.9, 1.0\}$, while dropout is set to 0.5 or 0.6. We set other parameters, such as the number of attention heads in the GAT layer and the embedding size, based on the specifications in the original paper for the backbone model.
We conducted each experiment 10 times with different random seeds and reported the average accuracy.
All experiments were conducted on Ubuntu 20.04 with NVIDIA RTX A6000 GPUs.

\subsection{Performance on Complex Biases (RQ1)}\label{section:main_exp}
To validate the effectiveness of our approach, we extensively compared CIE with state-of-the-art baselines on complex data biases, as detailed in Section \ref{section:exp_setting}.
Furthermore, we evaluated whether CIE is effective in low-resource scenarios by introducing the small sample selection bias.
Finally, we provided a comprehensive analysis and discussion of each experiment in Section \ref{section:Discussion_and_Analysis}.

\subsubsection{A Specific Type of Data Bias.}
To compare the performance of our proposed method with state-of-the-art methods under a specific type of bias, we separately introduce label selection bias and structural bias on three citation networks, and control the degree of each bias through $\epsilon$.
As shown in Table \ref{tab:label_selection} and \ref{tab:structural_bias}, we have the following findings:
\begin{itemize}
    \item Models that are tailored to address a specific type of data bias may exhibit suboptimal performance when deployed in other biases, occasionally even underperforming compared to GAT. Our empirical findings show that merely pursuing a specific type of data bias is inadequate to improve generalization performance overall, due to the challenge in anticipating the type of data biases in advance.
    \item In the experiments with a specific type of data bias, our proposed CIE significantly and consistently improves the performance of all backbone models, surpassing all state-of-the-art methods for debiased node classification.
    \item Our proposed method outperforms because the methods designed for specific types of bias lack global considerations. CIE considers the biases inherent in feature and structural information from a unified perspective, avoiding the challenge of designing for each bias separately.

\end{itemize}

% \usepackage{multirow}
\begin{table*}[]
\centering
\caption{Performance of label selection bias in three citation networks. The * indicates the best results of the baselines. Best results of all methods are indicated in bold. `\% gain over GNN' means the improvement percent against GNN (the same below).}

\resizebox{.95\textwidth}{!}{
        \begin{tabular}{c|cccc|cccc|cccc}
        \hline
        \multirow{2}{*}{Method} & \multicolumn{4}{c|}{\textbf{Cora}}                                    & \multicolumn{4}{c|}{\textbf{Citeseer}}                                & \multicolumn{4}{c}{\textbf{Pubmed}}                                       \\
                                & unbiased        & 0.4             & 0.6             & 0.8             & unbiased        & 0.4             & 0.6             & 0.8             & unbiased        & 0.4                 & 0.6             & 0.8             \\ \hline
        SGC                     & 0.7710          & 0.7723          & 0.7417          & 0.7330          & 0.6803          & 0.6717          & 0.6137          & 0.6480          & 0.7750          & 0.7477              & 0.7490          & 0.7517          \\
        APPNP                   & 0.8078          & 0.7808          & 0.7566          & 0.7336          & 0.6656          & 0.6474          & 0.6084          & 0.5842          & 0.7703          & 0.7650              & 0.7581          & 0.7574          \\
        CATs                    & 0.8093          & 0.7883          & 0.7560          & 0.7070          & 0.7017          & 0.7083          & 0.6593          & 0.6510          & 0.7770          & $\mathbf{0.7777^*}$ & 0.7613          & 0.7620          \\
        DGCN                    & 0.7947          & 0.7869          & 0.7653          & 0.7189          & 0.6734          & 0.6704          & 0.6349          & 0.6169          & 0.7849          & 0.7619              & 0.7680          & 0.7694          \\
        DGAT                    & $0.8141^*$      & $0.7950^*$      & $0.7775^*$      & $0.7579^*$      & $0.7070^*$      & $0.7132^*$      & $0.6861^*$      & $0.6705^*$      & 0.7812          & 0.7671              & $0.7717^*$      & 0.7713          \\
        EERM                    & 0.7974          & 0.7914          & 0.7698          & 0.7496          & 0.6986          & 0.7034          & 0.6764          & 0.6606          & 0.7754          & 0.7646              & 0.7675          & 0.7526          \\ \hline
        SAGE                    & 0.7822          & 0.7803          & 0.7377          & 0.7118          & 0.6824          & 0.6649          & 0.6723          & 0.6313          & 0.7721          & 0.7461              & 0.7424          & 0.7479          \\
        SAGE\_CIE               & 0.7944          & 0.7950          & 0.7537          & 0.7239          & 0.7038          & 0.6947          & 0.6975          & 0.6761          & 0.7741          & 0.7506              & 0.7534          & 0.7608          \\
        \% gain over SAGE       & 1.56\%          & 1.89\%          & 2.16\%          & 1.69\%          & 3.14\%          & 4.49\%          & 3.76\%          & 7.09\%          & 0.25\%          & 0.61\%              & 1.49\%          & 1.73\%          \\ \hline
        GCN                     & 0.7834          & 0.7777          & 0.7552          & 0.7101          & 0.6681          & 0.6634          & 0.6231          & 0.5877          & 0.7833          & 0.7536              & 0.7602          & $0.7714^*$      \\
        GCN\_CIE                & 0.8063          & 0.7999          & 0.7720          & 0.7285          & 0.6859          & 0.6822          & 0.6572          & 0.6231          & \textbf{0.7908} & 0.7689              & 0.7769          & 0.7776          \\
        \% gain over GCN        & 2.92\%          & 2.85\%          & 2.22\%          & 2.59\%          & 2.66\%          & 2.83\%          & 5.48\%          & 6.03\%          & 0.97\%          & 2.03\%              & 2.20\%          & 0.80\%          \\ \hline
        GAT                     & 0.8028          & 0.7850          & 0.7727          & 0.7381          & 0.7013          & 0.7047          & 0.6776          & 0.6544          & $0.7833^*$      & 0.7596              & 0.7668          & 0.7681          \\
        GAT\_CIE                & \textbf{0.8228} & \textbf{0.8047} & \textbf{0.7803} & \textbf{0.7627} & \textbf{0.7137} & \textbf{0.7236} & \textbf{0.7101} & \textbf{0.6799} & 0.7889          & 0.7721              & \textbf{0.7753} & \textbf{0.7763} \\
        \% gain over GAT        & 2.49\%          & 2.50\%          & 0.99\%          & 3.33\%          & 1.77\%          & 2.69\%          & 4.80\%          & 3.90\%          & 0.71\%          & 1.65\%              & 1.12\%          & 1.07\%          \\ \hline
        \end{tabular}
        }
        \label{tab:label_selection}
\end{table*}

% \usepackage{multirow}
\begin{table*}[]
\centering
\caption{Performance of structural bias in three citation networks. %The * indicates the best results of the baselines. Best results of all methods are indicated in bold. `\% gain over GNN' means the improvement percent of GNN\_CIE against GNN.
}

\resizebox{.95\textwidth}{!}{
        \begin{tabular}{c|cccc|cccc|cccc}
        \hline
        \multirow{2}{*}{Method} & \multicolumn{4}{c|}{\textbf{Cora}}                                    & \multicolumn{4}{c|}{\textbf{Citeseer}}                                & \multicolumn{4}{c}{\textbf{Pubmed}}                                       \\
                                & unbiased        & 0.4             & 0.6             & 0.8             & unbiased        & 0.4             & 0.6             & 0.8             & unbiased        & 0.4             & 0.6             & 0.8                 \\ \hline
        SGC                     & 0.7710          & 0.7640          & 0.7690          & 0.7707          & 0.6803          & 0.6917          & 0.6783          & 0.6894          & 0.7750          & 0.7653          & $0.7703^*$      & 0.7623              \\
        APPNP                   & 0.8078          & 0.7653          & 0.7462          & 0.7416          & 0.6656          & 0.6502          & 0.6258          & 0.6166          & 0.7703          & 0.7510          & 0.7661          & 0.7450              \\
        CATs                    & 0.8093          & $0.8043^*$      & $0.7930^*$      & $0.7760^*$      & 0.7017          & $0.6983^*$      & $0.6867^*$      & $0.6993^*$      & 0.7770          & $0.7707^*$      & 0.7677          & $\mathbf{0.7660^*}$ \\
        DGCN                    & 0.7947          & 0.7697          & 0.7524          & 0.7422          & 0.6734          & 0.6369          & 0.6407          & 0.6397          & 0.7849          & 0.7556          & 0.7594          & 0.7435              \\
        DGAT                    & $0.8141^*$      & 0.8014          & 0.7817          & 0.7701          & $0.7070^*$      & 0.6933          & 0.6837          & 0.6860          & 0.7812          & 0.7665          & 0.7647          & 0.7563              \\
        EERM                    & 0.7974          & 0.7858          & 0.7694          & 0.7638          & 0.6986          & 0.6806          & 0.6716          & 0.6808          & 0.7754          & 0.7682          & 0.7672          & 0.7468              \\ \hline
        SAGE                    & 0.7822          & 0.7474          & 0.6969          & 0.6007          & 0.6824          & 0.6557          & 0.6463          & 0.6439          & 0.7721          & 0.7488          & 0.7484          & 0.7293              \\
        SAGE\_CIE               & 0.7944          & 0.7664          & 0.7141          & 0.6502          & 0.7038          & 0.7008          & 0.6689          & 0.6803          & 0.7741          & 0.7603          & 0.7608          & 0.7365              \\
        \% gain over SAGE       & 1.56\%          & 2.54\%          & 2.46\%          & 8.25\%          & 3.14\%          & 6.87\%          & 3.50\%          & 5.65\%          & 0.25\%          & 1.54\%          & 1.66\%          & 0.98\%              \\ \hline
        GCN                     & 0.7834          & 0.7442          & 0.7246          & 0.6993          & 0.6681          & 0.6252          & 0.6278          & 0.6113          & 0.7833          & 0.7607          & 0.7587          & 0.7446              \\
        GCN\_CIE                & 0.8063          & 0.7784          & 0.7563          & 0.7444          & 0.6859          & 0.6565          & 0.6664          & 0.6610          & \textbf{0.7908} & 0.7619          & 0.7672          & 0.7556              \\
        \% gain over GCN        & 2.92\%          & 4.60\%          & 4.38\%          & 6.44\%          & 2.66\%          & 5.01\%          & 6.14\%          & 8.13\%          & 0.97\%          & 0.17\%          & 1.12\%          & 1.48\%              \\ \hline
        GAT                     & 0.8028          & 0.7833          & 0.7618          & 0.7429          & 0.7013          & 0.6885          & 0.6611          & 0.6737          & $0.7833^*$      & 0.7679          & 0.7684          & 0.7517              \\
        GAT\_CIE                & \textbf{0.8228} & \textbf{0.8109} & \textbf{0.7937} & \textbf{0.7834} & \textbf{0.7137} & \textbf{0.7024} & \textbf{0.6982} & \textbf{0.7062} & 0.7889          & \textbf{0.7783} & \textbf{0.7732} & 0.7561              \\
        \% gain over GAT        & 2.49\%          & 3.52\%          & 4.19\%          & 5.46\%          & 1.77\%          & 2.03\%          & 5.61\%          & 4.82\%          & 0.71\%          & 1.35\%          & 0.62\%          & 0.59\%              \\ \hline
        \end{tabular}
        }
        \label{tab:structural_bias}
\end{table*}

\subsubsection{Mixed Biases}\label{section:mixed_bias}
In real-world applications, models are often affected by multiple types of biases simultaneously, such as both label selection bias and structural bias, which is a more challenging and important issue.
To evaluate the performance of our proposed method under mixed biases, we introduce label selection bias and structural bias simultaneously on three citation networks as an illustration, and compare it with state-of-the-art methods.
As shown in Table \ref{tab:mixture}
(due to space limitations, the performance on the other two datasets is similar and can be found \href{https://github.com/Chen-GX/CIE}{here}.)
% (for detailed results on other datasets, please refer to Appendix \ref{section:appendix_mix_exp})
, we have the following findings:

\begin{itemize}
    \item Our proposed method significantly improves the generalization performance of the backbone model, even under mixed biases.
    By adopting a unified perspective on the distribution shift of feature and topology levels, CIE enables the model to effectively handle multiple types of biases simultaneously.
    \item Mixed biases have a more substantial influence on models than any specific type of bias. Nevertheless, in the experiments with mixed biases, the proposed method CIE consistently outperforms state-of-the-art debiased node classification methods by a considerable margin.
\end{itemize}

\begin{table*}[]
\centering
\caption{Performance of mixed biases (Label selection bias and Structural bias) on Cora. % The * indicates the best results of the baselines. Best results of all methods are indicated in bold. `\% gain over GNN' means the improvement percent of GNN\_CIE against GNN.
}
\resizebox{.8\textwidth}{!}{
\begin{tabular}{c|ccc|ccc|ccc}
\hline
\textbf{Label selection bias} & \multicolumn{3}{c|}{0.4}                            & \multicolumn{3}{c|}{0.6}                            & \multicolumn{3}{c}{0.8}                             \\
\textbf{Structural bias} & 0.4             & 0.6             & 0.8             & 0.4             & 0.6             & 0.8             & 0.4             & 0.6             & 0.8             \\ \hline
SGC                 & 0.7210          & 0.7100          & 0.6750          & 0.7240          & 0.6825          & 0.5925          & 0.7220          & 0.6800          & 0.6550          \\
APPNP               & 0.7430          & 0.6880          & 0.6520          & 0.7350          & 0.7400          & 0.6780          & 0.7190          & 0.6930          & 0.6780          \\
CAT                 & 0.7780          & 0.7505          & 0.7425          & 0.7255          & 0.7400          & 0.6935          & 0.7015          & 0.6885          & $0.6970^*$      \\
DGCN                & 0.7805          & 0.7165          & 0.7635          & 0.7195          & 0.7080          & 0.7125          & 0.6870          & 0.6970          & 0.6580          \\
DGAT                & $0.7900^*$      & $0.7560^*$      & $0.7670^*$      & $0.7420^*$      & 0.7115          & 0.6975          & $0.7220^*$      & 0.7220          & 0.6710          \\
EERM                & 0.7714          & 0.7363          & 0.7311          & 0.7368          & 0.7192          & 0.7012          & 0.7163          & 0.6906          & 0.6728          \\ \hline
SAGE                & 0.7650          & 0.7125          & 0.6435          & 0.6960          & 0.6655          & 0.5450          & 0.6370          & 0.6115          & 0.4660          \\
SAGE\_CIL           & 0.7795          & 0.7435          & 0.7070          & 0.7150          & 0.7190          & 0.6680          & 0.6935          & 0.6695          & 0.5650          \\
\% gain over SAGE   & 1.90\%          & 4.35\%          & 9.87\%          & 2.73\%          & 8.04\%          & 22.57\%         & 8.87\%          & 9.48\%          & 21.24\%         \\ \hline
GCN                 & 0.7693          & 0.7417          & 0.7253          & 0.7400          & 0.7113          & 0.6920          & 0.6893          & 0.7107          & 0.6303          \\
GCN\_CIL            & 0.7965          & \textbf{0.7595} & 0.7530          & 0.7590          & 0.7555          & 0.7100          & 0.7165          & 0.7255          & 0.6515          \\
\% gain over GCN    & 3.53\%          & 2.40\%          & 3.81\%          & 2.57\%          & 6.21\%          & 2.60\%          & 3.94\%          & 2.09\%          & 3.36\%          \\ \hline
GAT                 & 0.7835          & 0.7460          & 0.7585          & 0.7405          & $0.7565^*$      & $0.7360^*$      & 0.7200          & $0.7330^*$      & 0.6965          \\
GAT\_CIL            & \textbf{0.7965} & 0.7570          & \textbf{0.7900} & \textbf{0.7610} & \textbf{0.7575} & \textbf{0.7420} & \textbf{0.7565} & \textbf{0.7730} & \textbf{0.7210} \\
\% gain over GAT    & 1.66\%          & 1.47\%          & 4.15\%          & 2.77\%          & 0.13\%          & 0.82\%          & 5.07\%          & 5.46\%          & 3.52\%          \\ \hline
\end{tabular}
}
    \label{tab:mixture}
\end{table*}

\subsubsection{Evaluation in low-resource Scenarios.}
To evaluate the performance of CIE in low-resource scenarios, we introduce the small sample selection bias on NELL and Reddit datasets.
Due to the difficulty of most baselines in running on these datasets and not supporting neighbor sampling, we compare our method with MLP and backbone models using neighbor sampling. As shown in Table \ref{tab:small_sample}, we found that even in low-resource scenarios, our method can effectively improve the performance of backbone models. Furthermore, as the number of annotated nodes decreases, the improvement relative to the backbone model becomes more significant, further demonstrating the effectiveness of our method.

\begin{table*}[]
\centering
% \caption{Performance of small sample bias in NELL and Reddit. $\dag$ means GCN/GCN\_CIE for NELL and SAGE/SAGE\_CIE for Reddit. }
\caption{Performance of low-resource scenarios. % The * indicates the best results of the baselines. Best results of all methods are indicated in bold.
$\dag$ means GCN/GCN\_CIE for NELL and SAGE/SAGE\_CIE for Reddit. }

\resizebox{.8\textwidth}{!}{
        \begin{tabular}{c|cccc|cccc}
        \hline
        Method               & \textbf{NELL-1} & \textbf{NELL-3} & \textbf{NELL-5} & \textbf{NELL-10} & \textbf{Reddit-1} & \textbf{Reddit-3} & \textbf{Reddit-5} & \textbf{Reddit-10} \\ \hline
        MLP                  & 0.1816          & 0.3261          & 0.4685          & 0.5552           & 0.1359            & 0.1948            & 0.2433            & 0.2991             \\ \hline
        $\rm SAGE^\dag$      & 0.6233          & 0.6718          & 0.7812          & 0.7925           & 0.3558            & 0.5060            & 0.6722            & 0.7861             \\
        $\rm SAGE\_CIE^\dag$ & 0.6409          & 0.6945          & \textbf{0.7988} & 0.8019           & 0.4207            & 0.5480            & 0.7125            & 0.8145             \\
        \% gain over GCN     & 2.82\%          & 3.38\%          & 2.25\%          & 1.19\%           & 18.23\%           & 8.30\%            & 6.00\%            & 3.61\%             \\ \hline
        GAT                  & $0.6718^*$      & $0.7114^*$      & $0.7822^*$      & $0.8101^*$       & $0.7075^*$        & $0.8663^*$        & $0.8902^*$        & $0.9052^*$         \\
        GAT\_CIE             & \textbf{0.7079} & \textbf{0.7988} & 0.7977          & \textbf{0.8111}  & \textbf{0.7663}   & \textbf{0.8774}   & \textbf{0.9122}   & \textbf{0.9195}    \\
        \% gain over GAT     & 5.37\%          & 12.29\%         & 1.98\%          & 0.12\%           & 8.31\%            & 1.29\%            & 2.47\%            & 1.58\%             \\ \hline
        \end{tabular}
        }
        \label{tab:small_sample}
\end{table*}

\subsubsection{Discussion and Analysis}
\label{section:Discussion_and_Analysis}
Traditional node classification models neither distinguish between spurious features and causal features nor consider the susceptibility of spurious features to biases, which leads to performance degradation.
However, due to the selective aggregation of neighbor information (attention mechanisms), GAT generally outperforms other models such as GCN in most cases.
In debiased node classification methods, CATs are only designed to address structural bias, EERM requires sufficient labeled information to estimate and enrich the multiple environments, while DGNN is only designed to the feature-level distribution shift.
These methods are insufficient in addressing complex data biases, especially mixed biases.
DGNN estimates the weights of labeled nodes from a causal perspective, which in some cases can alleviate the interference of structural bias.
Therefore, DGNN achieves some favorable results compared to other baselines, but is still inferior to our proposed method.
Due to the requirement of a large amount of annotated information to generate multiple environments, EERM's performance in various semi-supervised and low-resource scenarios in this paper is not satisfactory and even inferior to models designed for specific biases.
For structural bias, we did not predict potential structures as EERM did because we found that this would introduce fatal erroneous information to the nodes with lower degrees.
Our proposed method takes a unified perspective on different types of biases: whether it will result in distribution shift at the feature level or topology level, both will affect the learning of node representations after information aggregation via GNNs.
Therefore, we constrain the model to learn causal features and mitigate spurious correlations at the node representation level.
This approach not only avoids the challenge of designing for each type of bias as in previous work, but also enables the model to handle multiple types of biases simultaneously under mixed biases.
Experimental results confirm the effectiveness of our method.

In summary, the above experimental results effectively validate the effectiveness of the proposed method under different types of biases and mixed biases. The proposed method CIE not only significantly improves the generalization performance of the backbone model but also surpasses state-of-the-art debiased node classification methods.

\subsection{Ablation Study and Sensitivity Analysis (RQ2 \& RQ3)}
\subsubsection{Ablation Study}
\begin{figure}[h]
    \centering
    \resizebox{0.7\linewidth}{!}{
    \includegraphics[width = \linewidth]{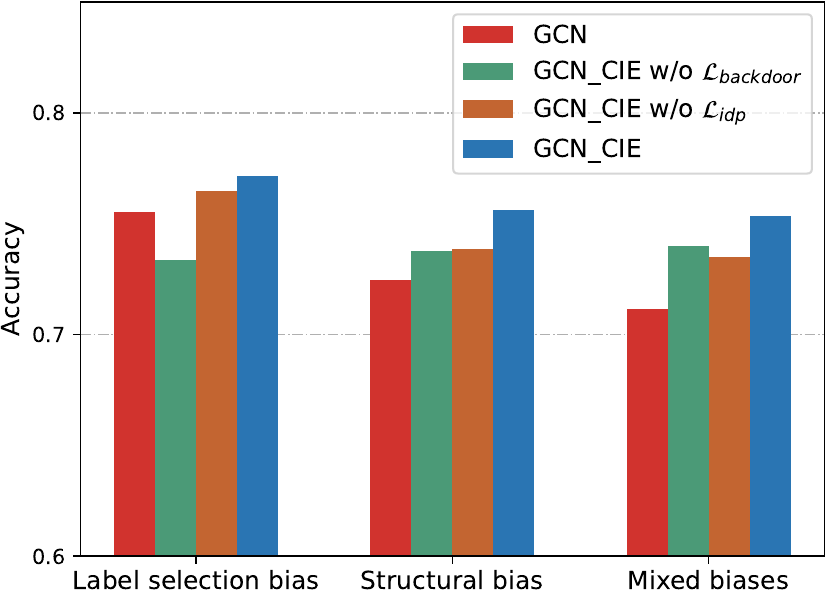}
    }
    \caption{Ablation Study. "w/o" means "without".
    % "w/o" is an abbreviation for "without".
    }
    \label{fig:ablation}
\end{figure}

In this section, we investigate the impact of the independence constraint and backdoor adjustment on overall performance through ablation experiments.
As shown in Table \ref{fig:ablation}, 
in most cases, using independence constraint or backdoor adjustment alone can improve generalization performance, but not stably. 
Intuitively, the independence constraint emphasize the independence of causal features and spurious features in subspaces, aiming to enhance the separation effect. The backdoor adjustment, on the other hand, aim to mitigate the impact of spurious correlation on the backdoor path.
The joint effect of independence constraint and backdoor adjustment can significantly improve the generalization performance of the backbone model while maintaining stability.

\subsubsection{Parameter Sensitivity.}
\begin{figure}[]
  \centering
  \begin{minipage}[t]{1\linewidth}
    \centering
    \includegraphics[width=\textwidth]{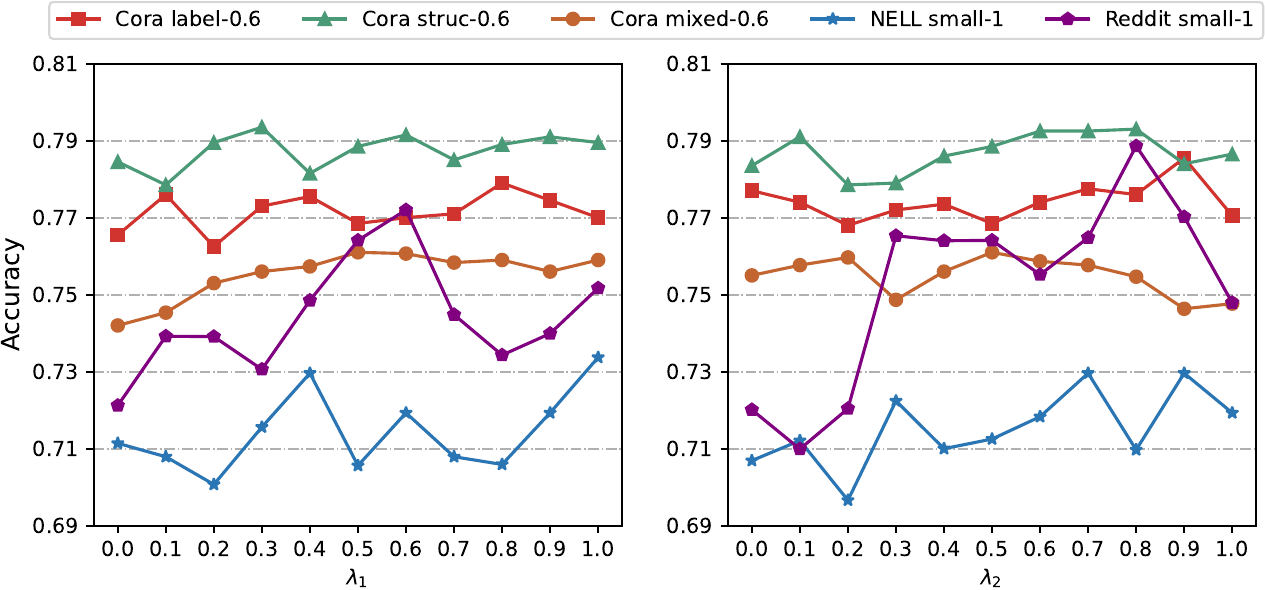}
    \caption{Parameter sensitivity of $\lambda_1$ and $\lambda_2$. "Cora label-0.6" denotes Label selection bias on Cora with $\epsilon$=0.6, "NELL small-1" denotes small sample selection bias on Nell with 1 samples per class.}
    \label{fig:sensitivity}
  \end{minipage}
  \qquad
  \begin{minipage}[t]{1\linewidth}
    \centering
    \includegraphics[width=\textwidth]{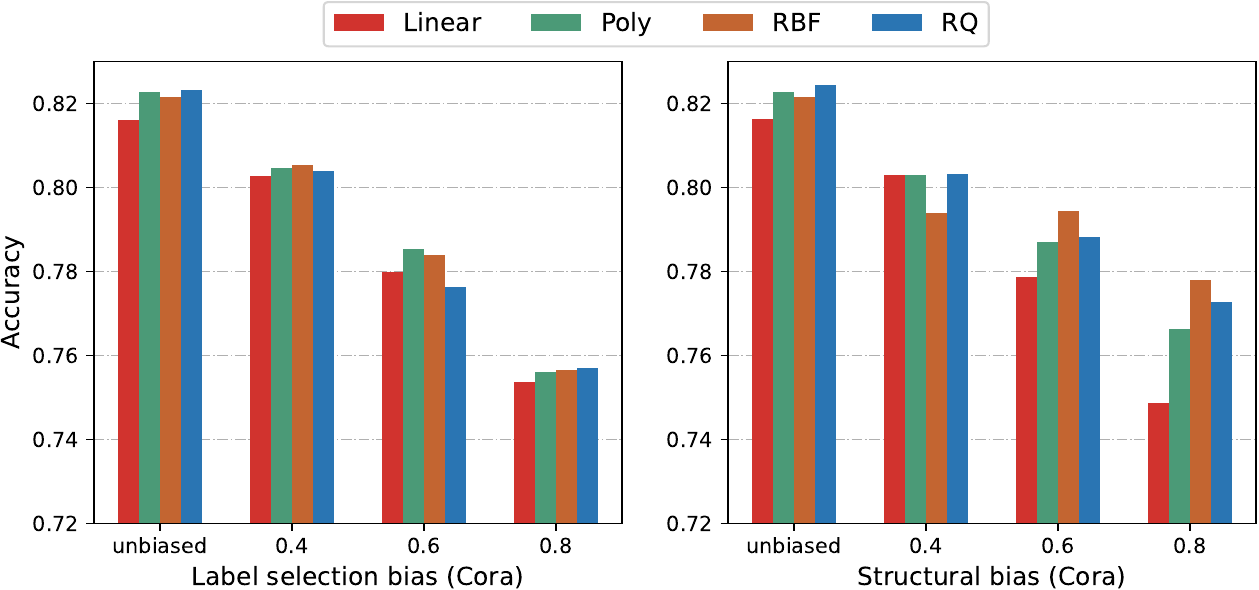}
    \caption{Sensitivity of the kernel function.}
    \label{fig:kernels}
  \end{minipage}
  % \caption{Overall caption of the figure}
  % \label{fig:main}
\end{figure}

We investigate the sensitivity of the parameters under different types of biases and report the results in Fig. \ref{fig:sensitivity}.
We select GAT as the backbone model, fix one parameter to 0.5, and the other parameter changes in [0,1] with a step size of 0.1.
The results show that the proposed method is relatively stable in most cases for the variation of $\lambda$, but fluctuates more in the small sample selection bias.
An intuitive explanation is that the limited amount of information provided when there are only a few samples per class makes it demand higher requirements for hyperparameters.
It is also worth noting that even in cases of fluctuations, there is a significant improvement in the performance of the backbone model, as shown in Table \ref{tab:small_sample} and Fig.\ref{fig:sensitivity}.

\subsubsection{Sensitivity of the kernel function.}
To investigate the sensitivity of the proposed method to the kernel function,
we fixed other parameters($\lambda_1=0.5,\lambda_2=0.5$), changed the linear kernel, polynomial kernel function (p=2, Poly), Radial Basis Function(RBF) and rational quadratic kernel (RQ), and the experimental results are shown in Fig. \ref{fig:kernels}.
In most cases, RBF and RQ are capable of achieving superior results.
It is evident that more complex kernel functions usually bring more significant improvements, especially when the model faces more complex biases.

\subsubsection{The computational complexity.}
We theoretically analyze the computational complexity of CIE.
The computational complexity of CIE is $\mathcal{O}(m^2)$, mainly from $\mathcal{L}_{idp}$, where $m$ is the number of training instances. 
The computational complexity of GNNs is mostly linear in the number of edges $|E|$ of the graph, e.g. the complexity of a single-layer GCN is $\mathcal{O}(|E|CHF)$ \cite{kipf2016semi}.
Therefore, the overall complexity of GCN\_CIE is $\mathcal{O}(|E|CHF + m^2)$.
We report the results of the average training time per epoch on different datasets, measured in seconds and shown in Table \ref{tab:time}.
Due to the typically small size of $m$ in semi-supervised node classification, the time complexity of CIE is on the same order of magnitude as the backbone model.
Moreover, CIE outperforms state-of-the-art debiased node classification models in terms of performance, while also exhibiting lower time complexity.

% \usepackage{booktabs}
\begin{table}[]
\caption{The training time per epoch (in seconds).}
\centering{
\resizebox{.75\linewidth}{!}{
    \begin{tabular}{cccc}
    \hline
             & Cora                  & Citeseer              & Pubmed                \\ \hline
    GCN      & $7.14\times 10^{-3}$  & $1.35\times 10^{-2}$  & $1.26 \times 10^{-2}$ \\
    GATs     & $4.87\times 10^{-2}$  & $5.08\times 10^{-2}$  & $5.11\times 10^{-2}$  \\
    DGCN     & $4.11 \times 10^{-2}$ & $5.35 \times 10^{-2}$ & $6.52 \times 10^{-2}$ \\
    EERM     & $5.23 \times 10^{-2}$ & $9.70 \times 10^{-2}$ & $1.84$                \\
    GCN\_CIE & $1.60\times 10^{-2}$  & $2.31 \times 10^{-2}$ & $2.74 \times 10^{-2}$ \\ \hline
    \end{tabular}
}
}
\label{tab:time}
\end{table}

\section{Conclusion} \label{section:conclusion}
In this paper, we propose a novel Causality and Independence Enhancement (CIE) framework, which takes a unified approach to handling multiple types of data biases simultaneously, eliminating the need for separately designed methods for each bias as in previous works. Specifically, it estimates causal and spurious features from node representations and mitigates spurious correlations through the backdoor adjustment. Furthermore, we introduce the independence constraint to enhance the discriminability and stability of causal and spurious features in complex biased environments. Extensive experiments demonstrate the effectiveness and flexibility of CIE. In future work, we aim to apply CIE to other tasks in different fields, such as out-of-distribution generalization in computer vision.

\begin{acks}
This work was funded by the National Key R\&D Program Young Scientists Project under grant number 2022YFB3102200 and the National Natural Science Foundation of China under grant number U21B2046. This work is also supported by the fellowship of China Postdoctoral Science Foundation 2022M713206.
\end{acks}

\bibliographystyle{ACM-Reference-Format} 
% \balance
\bibliography{sample-sigconf}

\end{document}